\documentclass[11pt,a4paper]{article}
\usepackage[hyperref]{emnlp2020}
\usepackage{times}
\usepackage{latexsym}

\usepackage{graphicx}
\usepackage{nccmath}
\usepackage{amsmath,amssymb}
\usepackage{xcolor}
\usepackage{longtable}
\usepackage[utf8]{inputenc}
\usepackage{booktabs}
\usepackage{bm}
\usepackage{array,multirow,graphicx}
\usepackage{float}
\usepackage{arydshln}
\usepackage{mathtools}
\usepackage{enumitem}
\usepackage{tikz}

\definecolor{ao-english}{rgb}{0.01, 0.28, 1.0}
\definecolor{awesome}{rgb}{1.0, 0.13, 0.32}
\definecolor{gray}{rgb}{0.5, 0.5, 0.5}

\usepackage[capitalize]{cleveref}
\usepackage{microtype}

\aclfinalcopy % Uncomment this line for the final submission

\title{Improving Zero Shot Learning Baselines with Commonsense Knowledge}
\author{Abhinaba Roy\(^\oslash\), Deepanway Ghosal\(^\dagger\), Erik Cambria\(^\oslash\),\\
\textbf{Navonil Majumder\(^\dagger\), \ Rada Mihalcea\(^\sqcap\) , \ Soujanya Poria\(^\dagger\) }\\ \\
\(^\oslash\)Nanyang Technological University, Singapore  \\
\(^\dagger\)Singapore University of Technology and Design, Singapore\\
\(^\sqcap\)University of Michigan, USA\\
\texttt{\{abhinaba.roy, cambria\}@ntu.edu.sg}, \\
\texttt{\{deepanway\_ghosal@mymail., sporia@, nmder@\}sutd.edu.sg},\\ 
\texttt{mihalcea@umich.edu}\\
}

\date{}

\begin{document}
\maketitle
\begin{abstract}
Zero shot learning --- the problem of training and testing on a completely disjoint set of classes --- relies greatly on its ability to transfer knowledge from train classes to test classes. Traditionally semantic embeddings consisting of human defined attributes (HA) or distributed word embeddings (DWE) are used to facilitate this transfer by improving the association between visual and semantic embeddings. In this paper, we take advantage of explicit relations between nodes defined in ConceptNet, a commonsense knowledge graph, to generate commonsense embeddings of the class labels by using a graph convolution network-based autoencoder. Our experiments performed on three standard benchmark datasets surpass the strong baselines when we fuse our commonsense embeddings with existing semantic embeddings i.e. HA and DWE.
\end{abstract}

% \section{Credits}

% This document has been adapted by Yulan He
% from the instructions for earlier ACL and NAACL proceedings, including those for 
% ACL 2020 by Steven Bethard, Ryan Cotterrell and Rui Yan, 
% ACL 2019 by Douwe Kiela and Ivan Vuli\'{c},
% NAACL 2019 by Stephanie Lukin and Alla Roskovskaya, 
% ACL 2018 by Shay Cohen, Kevin Gimpel, and Wei Lu, 
% NAACL 2018 by Margaret Michell and Stephanie Lukin,
% 2017/2018 (NA)ACL bibtex suggestions from Jason Eisner,
% ACL 2017 by Dan Gildea and Min-Yen Kan, 
% NAACL 2017 by Margaret Mitchell, 
% ACL 2012 by Maggie Li and Michael White, 
% ACL 2010 by Jing-Shing Chang and Philipp Koehn, 
% ACL 2008 by Johanna D. Moore, Simone Teufel, James Allan, and Sadaoki Furui, 
% ACL 2005 by Hwee Tou Ng and Kemal Oflazer, 
% ACL 2002 by Eugene Charniak and Dekang Lin, 
% and earlier ACL and EACL formats written by several people, including
% John Chen, Henry S. Thompson and Donald Walker.
% Additional elements were taken from the formatting instructions of the \emph{International Joint Conference on Artificial Intelligence} and the \emph{Conference on Computer Vision and Pattern Recognition}.

\section{Introduction}

\begin{figure}[t]
    \centering
    \includegraphics[width=0.8\linewidth]{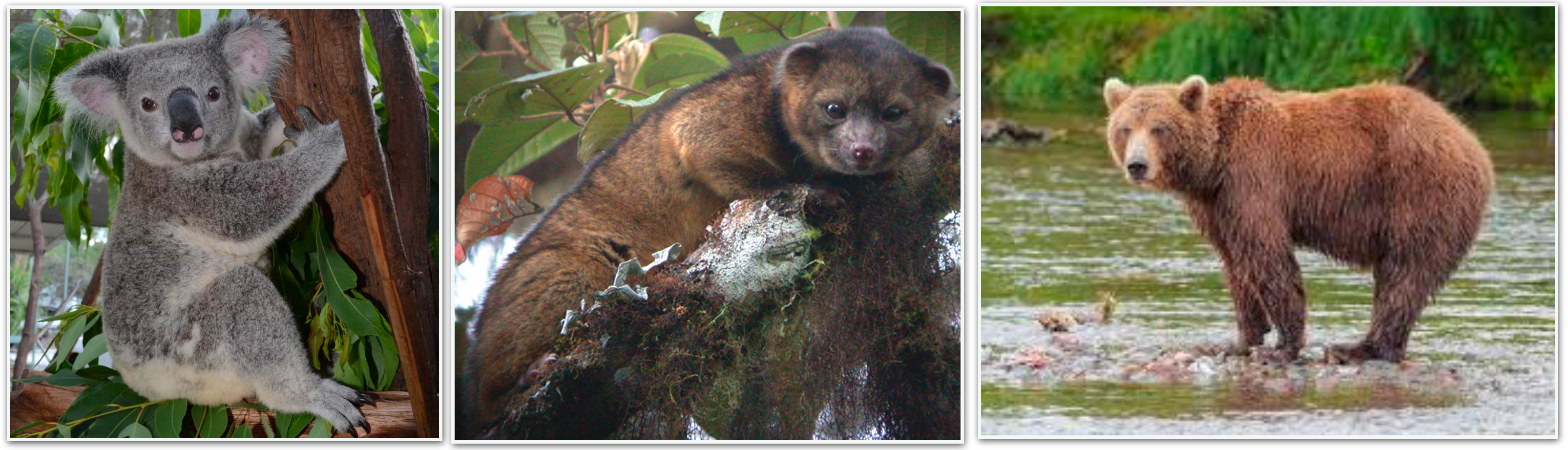}
    \caption{\footnotesize Animal in the middle is relatively unknown. However, we know the ones on the left and the right.
    % With the information "Olinguito has the \textbf{shape} of Koala and \textbf{color} of a Bear", it is easy to recognise the unknown animal correctly.
    }
    \vspace{-0.5cm}
    \label{fig:koala}
\end{figure}

How does one recognize an instance of an object they have never seen before? Instinctively, the first step is to find any resemblance with familiar objects. Take the example of \cref{fig:koala} --- with the information ``\emph{Olinguito has the shape of a Koala and the color of a Bear}'', we can establish shape- and color-based associations between known animals (Bear and Koala) and the unknown one (Olinguito). Inspired by such analogies, we explore how we can discover and exploit such associations between classes in zero-shot learning (ZSL) --- the problem of learning a visual classifier for a class with no training data. 

In a traditional ZSL setup, a visual classifier trained on a certain set of classes, namely \textit{seen} classes, is expected to perform reasonably well on a completely disjoint set of classes, namely \textit{unseen} classes. The key idea in dealing with such a scenario is to utilize knowledge from the \textit{seen} classes to characterize \textit{unseen} classes \cite{norouzi2013zero}. To this end, researchers often model the association between semantic embeddings of the classes and visual embeddings of the images. %The central idea of this association modeling is to learn a function using the training data that calculates association score. 
A high association score indicates a strong candidate class for the given image. During inference, each of the unseen class embeddings is paired with the visual embedding of the given image and fed to the association function. Typically, the class with the highest association score is selected as the predicted class. In literature, usually two types of semantic embeddings are used \cite{xian2017zero} --- human-defined attributes (HA) and distributed word embeddings (DWE). %Both of these mediums are conjoined together and called as \emph{semantic embeddings}.
% Zero-shot learning (ZSL) is the process of training a model on a set of classes (known classes) and testing the learnt model on completely different set of classes, known as unseen classes[ref][ref][ref]. A core hypothesis is in dealing with this situation is both seen and unseen classes share a set of common semantics in a higher dimensional space. In order to transfer knowledge from seen to unseen classes, we learn association between visual features and this semantic space and then generalize them over unseen classes. 

A number of approaches \cite{zhang2017learning,kodirov2017semantic,kumar2018generalized} use human-defined visually identifiable attributes (HA) such as \emph{stripes}, \emph{long\_neck}, \emph{furry}, \emph{spots} to describe classes. But, these approaches are rather expensive as human annotation is involved.

On the other hand, distributed word embeddings (DWE), such as word2vec %\footnote{\url{https://code.google.com/archive/p/word2vec/}}
successfully model different semantic properties of a language as a result of it being trained on a large-scale textual corpus \cite{jiang2017learning,kodirov2015unsupervised,qiao2017visually}. The primary objective of learning such word embeddings is based on the distributional hypothesis that states that similar words tend to appear in similar context. Using this hypothesis, pretrained word embeddings encode semantic properties of words. Nowadays, such embeddings are extensively used in many NLP tasks. However, these word embeddings are not trained on explicit knowledge representations to directly provide knowledge for an unseen word. These models can also struggle to find  similarity between the words that never or rarely appear in each other's context (see \cref{tab:analysis}).
%\hl{However, these embeddings encode contextual proximity between different word. This limits their ability to transfer visual similarity characteristics. }     
% \textit{\begin{figure}[t]
%     \centering
%     \includegraphics[width=\linewidth]{acl2020-templates/zsl_figures2}
%     \caption{\footnotesize Example of explicit association between two different classes. In aP\&Y dataset, class 'Airplane' is associated with class 'Motorbike' through Airplane \textcolor{red}{$\rightarrow$} \textit{Aeroplane} \textcolor{red}{$\rightarrow$} \textit{Aerodynamics} \textcolor{red}{$\rightarrow$} \textit{Motor racing} \textcolor{red}{$\rightarrow$} Motorbike. The relations (in red) that connect each of the nodes are explicitly defined in ConceptNet. Such external commonsense knowledge can be beneficial in ZSL.}
%     \label{fig:plane_bike}
% \end{figure}

\begin{figure}[H]
    \centering
    \includegraphics[width=0.8\linewidth]{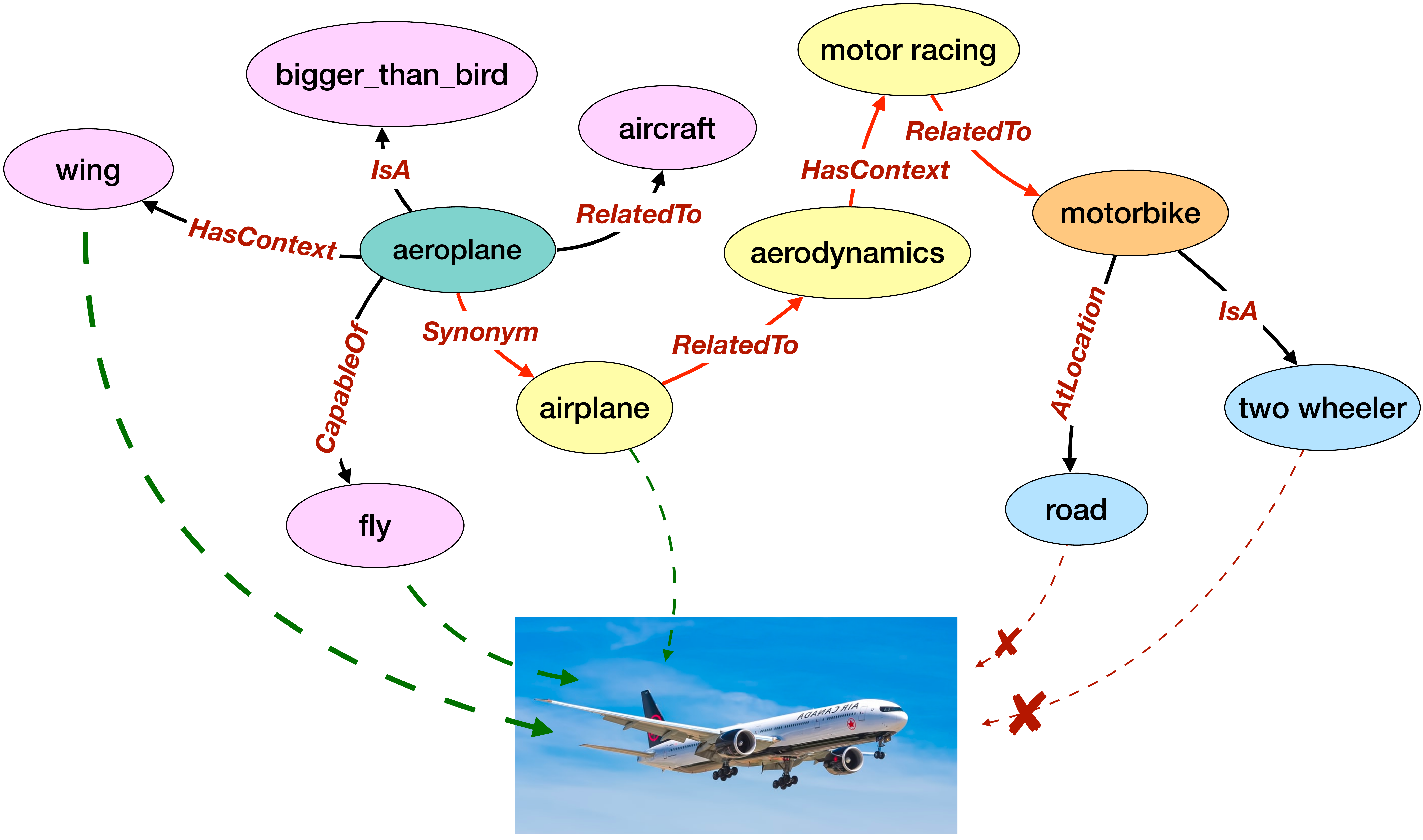}
    % \caption{Caption}
    % 
    \caption{An example of commonsense relations between \textit{Aeroplane} and \textit{Motorbike} in ConceptNet.}
    \vspace{-0.2cm}
    \label{fig:cnet_relation}
\end{figure}

%Most of the research works on ZSL try improving the association between visual and semantic embeddings \cite{zhang2017learning,sung2018learning}. According to these works, boosting this association helps in transferring knowledge from seen to unseen classes.
Returning to the example in \cref{fig:koala}, we can observe that external knowledge helps in finding an explicit relation between different classes by traversing through the paths between them in the graph. To elucidate this, consider the example in \cref{fig:cnet_relation} ---  class `Aeroplane' (seen class) is connected to class `Motorbike' through `Motor racing' (unseen class). In this example, the relations (in red) that connect the nodes (in yellow) on the path from \emph{aeroplane} to \emph{motorbike} are explicitly defined in ConceptNet \cite{speer2017conceptnet}. Such a relational knowledge-graph that contains relations among the words can efficiently model associations that can be very beneficial to the ZSL task by transferring knowledge from seen to unseen classes. %An elucidated example is shown in Figure~\ref{fig:plane_bike}. 
% \noindent \resizebox{\linewidth}{!}{$aeroplane\xrightarrow{\color{red}{Synonym}}airplane\xrightarrow{\color{red}{RelatedTo}} 
% aerodynamics\xrightarrow{\color{red}{HasContext}}motorracing\xrightarrow{\color{red}{RelatedTo}}motorbike$} 
Another benefit of using knowledge graphs for ZSL is that it can provide additional background information on the class labels, which in turn can improve the semantic embeddings. An illustration in \cref{fig:cnet_relation} shows the background knowledge of \emph{aeroplane} being represented with the nodes in magenta. 
% e.g., \emph{$\text{aeroplane}\xrightarrow{CapableOf}\text{fly}$}. Even someone who did not previously know the term \emph{aeroplane} should be able to make inferences about its meaning after seeing this background knowledge. We therefore hypothesize that improvements in the semantic embeddings will also lead to improvements in the association of visual and semantic embeddings.
For someone oblivious to an aeroplane, the shape and size of it is still imaginable through the associated background commonsense knowledge. For instance, $aeroplane\xrightarrow{\textcolor{red}{CapableOf}}\text{fly}$ or $aeroplane\xrightarrow{\textcolor{red}{HasContext}}\text{wing}$ are strong conceptual indicators to aeroplanes, but not to motorbikes. Naturally, the semantic embeddings, created from such knowledge graph, will be vastly different for an aeroplane as compared to that of a motorbike, leading to stronger association between the representation of the input image in \cref{fig:cnet_relation} and aeroplane than motorbike. We therefore hypothesize that the improved semantic embeddings will lead to improved association of visual and semantic embeddings, resulting to correct class prediction during inference. We achieve this by fusing commonsense embeddings, extracted from ConceptNet, with the human-defined attributes (HA) and distributed word embeddings (DWE).

% ConceptNet is one such large-scale knowledge graph that contains commonsense knowledge (c.f. \cref{fig:cnet_relation}).
\emph{Commonsense knowledge} broadly represents the large amount of small but well-known facts that we often take for granted \cite{andrich2009common}, and yet we regularly use to understand our surrounding context or to interpret or synthesize language. To leverage commonsense-knowledge, we employ ConceptNet~\cite{speer2017conceptnet}, a  large-scale knowledge graph that contains commonsense knowledge. We posit that using such commonsense knowledge is beneficial for our task, as compared to other knowledge graphs, such as WordNet which only contain relations like \emph{IsA}, \emph{Synonymy}, \emph{Hypernymy}. %etc. In Figure \ref{fig:plane_bike}, we observe a diversified set of relations from ConceptNet connecting `aeroplane' and `motorbike'. These relations are explicit and clearly defined, which eventually makes knowledge transfer easy from \textit{seen} classes to \textit{unseen} classes.
 
We first train a global (covering all possible classes as nodes) graph autoencoder on ConceptNet to learn inter-class associations. To this end, a graph convolutional network (GCN) is employed for relation prediction in the graph. The learnt node embeddings are used as the commonsense embeddings (CSE) of the corresponding class names.
Further, we fuse this CSE with the semantic embeddings.
% ---  human-defined attributes and distributed word embeddings.
We surmise that the obtained CSE encode associations between \textit{seen} and \textit{unseen} classes in the ZSL setup. Additionally, CSE should also encode key background information about the image class labels. To validate our claim, we use the CSE of the class labels to induce external commonsense knowledge in two different ZSL baselines. We observe noteworthy improvements in ZSL performance throughout our experiments across three different benchmark ZSL datasets. %\hl{For example, there is an improvement of 65.35\% when CSE is infused with word2vec for AwA2 dataset.}
% These strategies aid the ZSL baseline models to discover associations between \textit{seen} and \textit{unseen} classes. 

% \hl{motivate with an example why commonsense. Why do you think external knowledge can help. What is the different with Glove.}

The main contributions of this paper are --
\begin{itemize}[leftmargin =*,noitemsep]
    \item We encode the commonsense knowledge in ConceptNet using a graph convolutional network (GCN)-based autoencoder; %The extracted commonsense embeddings (CSE) from this process improve ZSL performance.
    \item We improve association between visual and semantic embeddings by fusing existing semantic embeddings (HA and DWE) with commonsense embeddings (CSE); 
    \item We demonstrate effectiveness of CSE through extensive experiments and comparisons with two different ZSL baseline methods across three different datasets.
\end{itemize}

The rest of this paper is organized as follows: \S \ref{sec:relatedwork} briefly reviews ZSL approaches related to this work; \S \ref{sec:background} provides task definition and the details of the two baselines that we use in this work; \S \ref{sec:method} describes our method; we report findings of our experiments and related discussions in \S \ref{sec:experiments}; finally, \S \ref{sec:conclusion} concludes this paper.
% in Attribute based methods,\hl{give citation} human define \textit{attributes} describe visual feature identified and quantified by annotators. Generally, these are binary values that signify presence or absence of a certain visual characterization. Each class is defined as a certain combination of these attributes and knowledge transfer is done based share-ability of them. For example, if a model is trained on cheetah, it can also be transferred to classify leopard based on the prior knowledge that leopard share similar shape attributes with cheetah but has different sized spots in its body. %As both local and deep features extracted from images focus on shape information, human defined visual attributes align well with them provide good performance in Zeor-shot image recognition.%
% But visual attributes are rather expensive to train as manual inspection and annotation is involved. An alternative and cheaper option is to use distributed word embedding(DWE)s such as word2vec[ref] and GloVe [ref]. They are trained on a large text corpus and produce similar embeddings for words appearing in similar context. Going back to the example of Cheetah and 

\section{Related Work}
\label{sec:relatedwork}
% Incorporating novel classes has also gained notable attention.
Early works on zero-shot learning use human defined attributes \cite{farhadi2009describing,AwA1,jayaraman2014zero} to represent each class as a vector that denotes presence/absence of attributes. Direct mapping between image features and these class vectors is used to learn visual classifiers.

Another line of work use distributed word embeddings (DWE)s such as word2vec \cite{mikolov2013distributed} to represent each category. These are consecutively mapped to visual classifiers \cite{frome2013devise,norouzi2013zero,kodirov2015unsupervised, roy2018visually, qiao2017visually}. \citet{frome2013devise} use convolutional neural network (CNN) and a transformation layer to train a mapping from image to word embeddings. %Alternatively, ConSE \citep{norouzi2013zero} constructs the image embedding by combining an existing image classification CNN and DWE model. 
Few recent works \cite{kodirov2015unsupervised, roy2018visually, qiao2017visually} propose to reduce the gap between CNN image features and DWEs. Combination of attributes and distributed word embeddings are used in works such as \cite{akata2013label,zhang2017learning}. %Wikipedia text has been used as a wealthy source of information in \cite{lei2015predicting,qiao2016less}.

Our work is also related to recent approaches that use knowledge graphs to distill knowledge explicitly from knowledge representations \cite{salakhutdinov2011learning,deng2014large,kampffmeyer2019rethinking,wang2018zero}. Many approaches exist in object recognition that use such approach. \cite{salakhutdinov2011learning} uses a hierarchical classification model that allows rare objects to borrow statistical strength from related objects that have many training examples. %\cite{deng2014large} uses a model that allows encoding of flexible relations between labels. They apply certain constraints during training of object classifiers.  \cite{kampffmeyer2019rethinking} Proposes Dense Graph Propagation (DGP) module to resolve the problem of vanishing information through distant nodes.

Closest to our work is \cite{wang2018zero}. Although our approach is similar to theirs, in that we both use Graph Convolution Network (GCN) to improve class embeddings, the way we use GCN is vastly different. %They use GCN on a knowledge graph to transform initial pre-trained word embeddings and find an association among the transformed embeddings to visual embedding. However, we extract sub-graphs for all seen and unseen classes from the knowledge graph and pass them through a graph convolutional autoencoder model to extract a commonsense embedding.

% They use GCN on NELL to transform initial pre-trained word embeddings and map those transformed embeddings to visual embedding directly. However, we extract sub-graphs for all seen and unseen classes from ConceptNet and pass them through a graph convolutional autoencoder model to extract a commonsense embedding for the same. We fuse them with existing semantic embeddings to be used in the ZSL framework. %Technical details of our method is provided in the following section.
\begin{figure*}[t]
    \centering
    \includegraphics[width=0.8\linewidth]{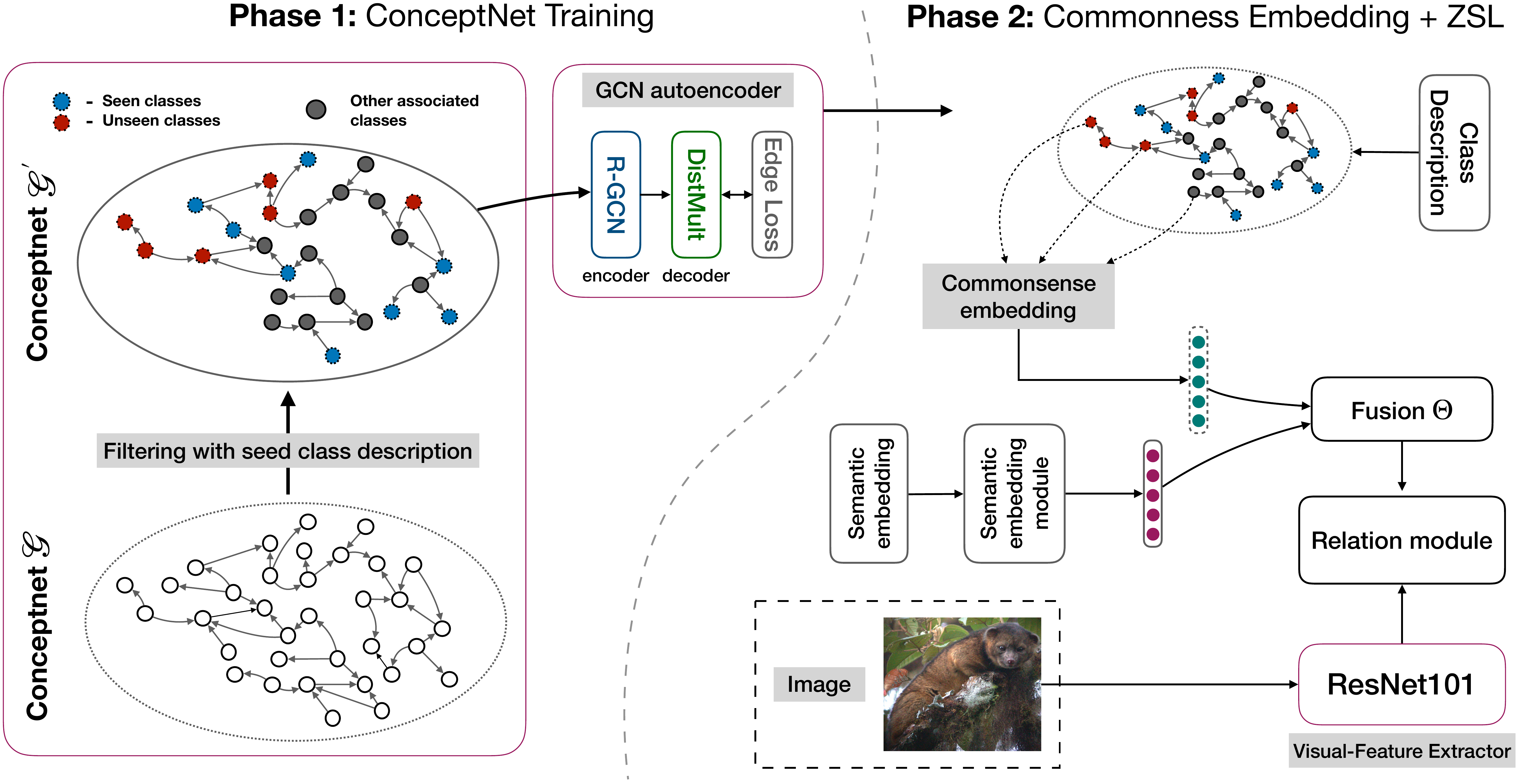}
    \caption{\footnotesize The overall idea of our proposed framework.}
    
     \vspace{-0.2cm}
    % \label{fig:koala}
\end{figure*}
\section{Background}

\label{sec:background}
%In this section, we provide a brief definition of the task of Zero Shot Learning (ZSL) and describe the baseline frameworks in our experimental studies.
\subsection{Task Definition}
\label{sec:prelem}
%\hl{define semantic space
In a ZSL setup, we have a set of training instances (in this case, these are images), $X_{train}$ belonging to a set of classes $\mathcal{Y}_{seen}$. We test on a set of images, $X_{test}$ belonging to a set of classes $\mathcal{Y}_{unseen}$. It is assumed that $\mathcal{Y}_{seen} \cap \mathcal{Y}_{unseen} = \emptyset$. Ideally, for each $x_i \in X_{train}$, we extract its visual embedding $f(x_i) \in \mathbb{R}^{m}$ by passing it through a visual feature extractor such as a deep CNN \citep{simonyan2014very}. Each class $y \in \mathcal{Y}_{seen \cup unseen}$ has a semantic embedding $s_{y} \in \mathbb{R}^{N}$. $s_y$ can be human defined attributes of the image class-label such as \emph{stripes, furry} or distributed word embeddings of the image class (such as word2vec).
In a standard ZSL training, we use $\mathcal{Y}_{seen}$ class data to learn relation module relation between $f(x) ; \forall x \in X_{train}$ and $s_y; \forall y \in \mathcal{Y}_{train}$. 
During testing, we use the same relation module to find the corresponding $\widetilde{y} \in \mathcal{Y}_{unseen}$ for a test image $\widetilde{x} \in X_{test}$. 
%Success of a ZSL approach depends on how $\Theta$ is defined as well as choice of semantic embedding. 
\subsection{Baseline Methods}
We utilize the commonsense knowledge extracted from ConceptNet in two different strong ZSL baselines: 1) Deep Embedding ZSL; 2) Relation Network.
% \begin{enumerate}[leftmargin=*]
% \item Deep Embedding ZSL,
% \item Relation Networks. 
% \end{enumerate}
Each of them consists of two main modules:
\begin{itemize}[leftmargin=*,noitemsep]
    \item \textbf{Semantic embedding module} encodes the semantic embedding of the image class-label embedding of the image. We refer to this module as the $\Theta$ network.
    \item \textbf{Relation module} takes input from the semantic embedding module and visual embeddings. It aims to model the association between the semantic embedding of the image class-label and the visual/image embedding.
\end{itemize}
We describe the two baselines in more detail below.
\subsubsection{Deep Embedding ZSL (DeZSL)}
\label{sec:baseline1}
%%%%%%%%%%%%% Navonil %%%%%%%%%%%%%%%%%%
Following the work of \citet{zhang2017learning}, we jointly encode both image and semantic
representations into a unified embedding space.
\paragraph{Semantic embedding module:} The $m$-dimensional semantic embeddings of the class-label $y$, namely $s_y$, is fed to the semantic embedding module named the $\Theta$ network, composed of two fully-connected (FC) layers with ReLU
activation, for projection onto visual embedding space:
\begin{flalign}
        \Theta_{DeZSL} &= FC^{Relu}_{b}(FC^{Relu}_{a})
\end{flalign}The output size of the $FC^{Relu}_{b}$ is kept equal to image feature dimension.
%%%%%%%%%%%%%%%%%%%%%%%%%%%%%%%%%%%%%%%%
%We motivate our first baseline from the work of \cite{zhang2017learning}. This uses visual feature space of a deep neural network (DNN) as a common embedding space for both images and semantic representation of class that the image belongs to.\\

%\noindent \textbf{Embedding module}: Embedding module of the DeZSL model takes $m$ dimensional semantic representations which can either be the manually labelled attribute features or the distributed word embedding of the \emph{image class} and pass it through a shallow network $\Theta$ to make the embedding of equal length to the visual feature (\hl{Figure ?}). This function is composed of two fully connected (FC) layers with ReLU activation (\hl{Figure 3}). So, in this case: \begin{equation}
 %       \Theta = FC^{Relu}_{B}(FC^{Relu}_{A})
%    \end{equation}where $FC^{Relu}{}$ is a fully connected (FC) network with ReLU activation. The output size of the $FC^{Relu}_{b}$ is kept equal to image feature dimension.

\noindent \textbf{Relation module}: The relation module tries to minimize distance between the image feature $f(x_i)$ of an image $x_i$ and the corresponding class semantic representation $\Theta(s_{y_i})$ ($y_i$ is the class-label of $x_i$). \cref{eqn:baseline1obj} shows the related objective function. 
\begin{equation}
    \begin{aligned} 
    \mathcal{L}=\frac{1}{N} \sum_{i=1}^{N} \| f(x_i)- \Theta_{DeZSL}(s_{y_i})\|^{2}
    \end{aligned}
    \label{eqn:baseline1obj}
\end{equation}

%We design two different $\Theta$ for two possible different scenarios.
%\begin{enumerate}
%    \item 
 
%\item When fusing two different semantic spaces $s_1$ and $s_2$ (as in Figure 3), we first map each of them to a fusion layer where they are concatenated. This is followed by FC+ReLU unit and is projected them onto visual space(Figure 4). 
 %   \begin{equation}
%        \begin{aligned}
    %\begin{alignat*}
 %      &\Theta_1 = g(h(s_1) \\
 %      &\Theta_2 = g(h(s_2) \\
 %       &\Theta = g(h(\Theta_1\bm{+}\Theta_2))
       % \end{alignat*}
%     \end{aligned}
 %   \end{equation}
     
%\end{enumerate} 

\noindent \textbf{Inference}: For an unseen image $\widetilde{x}$, we find the nearest of all possible prototype $\Theta(s);  \forall s \in \mathcal{Y}_{unseen}$ in visual embedding space, denoted as:
\begin{equation}
\begin{aligned}
     y = \arg \min _{y} \mathcal{D}(f(\widetilde{x}),\Theta_{DeZSL}(s_{y}))%;\\ \forall y \in %\mathcal{Y}_{test} 
\end{aligned}
\end{equation}
Here, $\mathcal{D}$ is a distance function and $y \in \mathcal{Y}_{test}$.

\subsubsection{Relation Network (RN)}
\label{sec:baseline2}
We use a two-branch Relation Network following the work of \cite{sung2018learning} as our second baseline approach. \\

\noindent \textbf{Semantic embedding module}: It takes semantic representations of the image class-label $s_y$ as inputs, passes it to $\Theta$ and produces a representation which later is sent to the relation module to model the relation and association between the image class-label and the image features.
\begin{equation}
    \Theta_{RN} = FC^{Relu}_{c}
\end{equation}
Finally we obtain $\Theta_{RN}(s_y)$ where $y \in \mathcal{Y}_{train}$

\noindent \textbf{Relation module}: Relation module of the RN takes concatenated image and semantic representations of the image class-label and passes them through a non-linear transformation using $FC^{Relu}$ layer to model the association and relation among these two different types of representations.  
\begin{equation}
        \Gamma(s_y, x) = FC^{Relu}_{d}(\Theta_{RN}(s_y) \oplus f(x))
        \label{eqn:baseline2gamma}
\end{equation}
where $x \in X_{train}$, $f(x)$ represents visual features and $\oplus$ is the concatenation operation. In order to calculate a single score to measure the relation and association between the image class-label embeddings and the image features, $\Gamma(s_y, x)$ is fed to a FC layer of output size M with sigmoid activation where M is the number of unique image class-labels in the training data.

\begin{equation}
        R(s_{y}, x) = FC_{sigmoid}(\Gamma(s_y, x))
        \label{eqn:rfunction}
\end{equation}
where $s_y$ is the set of all unique class-labels i.e., ${s_{y_1}, s_{y_2}, s_{y_3},.....,s_{y_M}}$ in the training dataset.

Finally the model is trained using MSE loss using the following equation:
 \begin{equation}
 \small
     \begin{aligned} 
    \mathcal{L}=\frac{1}{N} \sum_{i=1}^{N} \sum_{j=1}^{M} \| R(x_i, s_{y_j})- \textbf{1}(s_{y_i} == s_{y_j})\|^{2}
     \end{aligned}
    \label{eqn:baseline1}
\end{equation}

\noindent \textbf{Inference}: For an unseen image $\widetilde{x} \in X_{test}$, we build the relation pairs with all the unique image class-labels and choose the pair with the highest value of $R(s_y, \widetilde{x})$  according to \cref{eqn:rfunction}. 
\begin{equation}
    y = \arg \max _{y}(R(s_y, \widetilde{x}))
\end{equation}
where $y \in \mathcal{Y}_{test}$.

We incorporate our commonsense knowledge embedding with attribute/semantic embeddings in each of the $\Theta$s to improve the existing semantic embedding (human defined attribute or woprd2vec embedding) $s_y$ in our experiments.

\section{Method}
\label{sec:method}
Our method integrates commonsense knowledge extracted from ConceptNet, and leverages commonsense embeddings obtained from graph convolutional autoencoders to improve ZSL. Formally, the method consists of two sequential phases:
\paragraph{Phase 1:} This phase deals with training a class-aggregated sub-graph of ConceptNet. In particular: a) We create a sub-graph of ConceptNet based on all seen-unseen classes (\S \ref{sec:subgraph_construction}), and b) We train a graph-convolutional autoencoder model to learn CSE of the classes
% ~\cite{DBLP:conf/esws/SchlichtkrullKB18}
(\S \ref{sec:kb_training}).
\paragraph{Phase 2:} After the graph autoencoder model is trained,
% ~\footnote{We use \textit{node}, \textit{concept}, and \textit{entity} interchangeably.}, 
a) We extract CSE for each of the classes from the graph (\S \ref{sec:kb_feature_extract}), and b) We feed the corresponding CSE  into our ZSL framework (\S \ref{sec:baseline1},\S \ref{sec:baseline2}).

\noindent 
We define each of these steps in detail below:

\subsection{Phase 1: a) Class-Aggregated Commonsense Graph Construction}
\label{sec:subgraph_construction}

We construct our class-aggregated graph from ConceptNet~\cite{speer2017conceptnet}. First, we introduce the following notation: ConceptNet graph is represented as a directed labeled graph $\mathcal{G = (V, E, R)}$, with concepts/nodes $v_{i} \in \mathcal{V}$ and labeled edges $(v_i, r_{ij}, v_j)\in \mathcal{E}$, where $r_{ij} \in \mathcal{R}$ is the relation type of the edge between $v_i$ and $v_j$. %Since the graph is directed, two concepts can have edges in both directions with different relations. 
The concepts in ConceptNet are unigram words or n-gram phrases. For instance, a labelled edge from ConceptNet is 
(\textit{baking-oven}, \textit{\textcolor{red}{AtLocation}}, \textit{kitchen}), where baking-oven and kitchen are concept nodes and  \textit{\textcolor{red}{AtLocation}} is the relation.

ConceptNet has approximately 34 million edges, from which we first extract a subset of edges based on the following heuristic. The seen and unseen class names are treated as the seeds that we use to filter ConceptNet into a sub-graph. In particular, we extract all the triplets from $\mathcal{G}$ which are within the vicinity of radius 2 of any of those seed class names, resulting in a sub-graph $\mathcal{G'} = (V', E', R')$, with approximately 500K edges. 

% We surmise that this scheme of creating $\mathcal{G'}$ would result in a sub-graph with inter-class associations across all seen and unseen classes. 
% The sub-graph of the words common for all the domains and inter-domain bridges (\cref{fig:conceptual_links}) will ideally capture the domain-general knowledge, which will be transferred during domain adaptation. Furthermore, neighbours of the sentiment bearing words distinct in each domain will help in learning better representations for domain-specific knowledge required for the classification.
\subsection{Phase 1: b) Knowledge Graph Pre-training}
\label{sec:kb_training}
To utilize $\mathcal{G'}$ in our task, we first need to compute a representation of its nodes. We do this by training a graph autoencoder model to perform link prediction
% --prediction of new facts--which is 
(a standard knowledge graph completion task). The model takes as input an incomplete set of edges $\hat{\mathcal{E'}}$ from $\mathcal{E'}$ in $\mathcal{G'}$ and then assign scores to possible edges $(c_1, r, c_2)$, determining how likely are these edges to be in $\mathcal{E'}$.

Following \citet{DBLP:conf/esws/SchlichtkrullKB18}, our graph autoencoder model has two modules: a R-GCN entity encoder and a DistMult scoring decoder.

\paragraph{Encoder Module.}
We employ the Relational Graph Convolutional Network (R-GCN) encoder from  \citet{DBLP:conf/esws/SchlichtkrullKB18} as our graph encoder network. The R-GCN transformation encodes concepts in a multi-relational graph and can be understood as a special case of a basic differentiable message passing~\cite{gilmer2017neural}. The power of this model comes from its ability to accumulate relational evidence in multiple inference steps from the local neighbourhood around a given concept. The neighborhood-based convolutional feature transformation process always ensures that distinct classes are connected to each other via latent concepts and influence each other to create enriched class-aggregated feature vectors.  

Specifically, our encoder module consists of two R-GCN encoders stacked upon one another. This enables information flow from concepts that are within a distance of two hops. Extending this by the rule of associativity ($a\rightarrow b\rightarrow c$ and $c\leftarrow d\leftarrow e$) the model can capture a relatively long path relation ($a\rightarrow b\rightarrow c\leftarrow d\leftarrow e$). An alternate way to make the model capable of capturing long path relations would be to make the encoder module deeper. However, stacking more layers results in over-smoothing as node features eventually converge to the same value. Training deep GCN networks is thus very difficult and still an open research problem.

The initial concept feature vector $g_i$ is initialized randomly and thereafter transformed into the class-aggregated feature vector $h_{i} \in \mathbb{R}^d$ using the two-step graph convolution process. The transformation process is detailed below:
\begin{equation}
\small
\begin{aligned}
    & f(x_i , l)   = \sigma (\sum\limits_{r \in \mathcal{R}}^{} \sum\limits_{j \in N_{i}^{r}}^{} \frac{1}{c_{i,r}}W_{r}^{(l)} x_{j} + W_{0}^{(l)}x_{i}) \\
% \end{aligned}
% \end{equation}
% \begin{equation}
% \small
% \begin{aligned}
   & h_{i} = h_{i}^{(2)}  = f(h_i^{(1)},2) \quad ; \quad h_{i}^{(1)} = f(g_i \, 1) \label{eq-ref1}
\end{aligned}
\end{equation}
% \begin{align*}
% \small
%     f(\mathbf{x}_i , l)   = \sigma (\sum\limits_{r \in \mathcal{R}}^{} \sum\limits_{j \in N_{i}^{r}}^{} \frac{1}{c_{i,r}}W_{r}^{(l)} \mathbf{x}_{j} + W_{0}^{(l)}\mathbf{x}_{i}) \\
%     \mathbf{h}_{i} = \mathbf{h}_{i}^{(2)}  = f(\mathbf{h}_i^{(1)},2) \quad ; \quad \mathbf{h}_{i}^{(1)} = f(\mathbf{g}_i \, 1)
% \end{align*}
where, $N_{i}^{r}$ denotes the neighbouring concepts of concept $i$ under relation $r \in \mathcal{R}$; $c_{i,r}$ is a problem specific normalization constant which can be set in advance, such that, $c_{i,r} = |N_{i}^{r}|$, or can be learned in a gradient-based learning setup. $\sigma$ is an activation function such as ReLU, and $W_{r}^{(1/2)}$, $W_{0}^{(1/2)}$ are learnable parameters of the transformation.

This stack of transformations (\cref{eq-ref1}) effectively accumulates normalized sum of the local neighbourhood 
% (features of the neighbours) 
i.e. the neighbourhood information for each concept in the graph. The self connection ensures self-dependent feature transformation.

\paragraph{Decoder Module.}
DistMult factorization \cite{yang2014embedding} is used to score a triplet $(c_i, r, c_j)$,
\begin{align}
    \begin{split}
        s(c_i, r, c_j) = \sigma(h_{c_i}^{T}R_{r}h_{c_j})
        \label{eq-ref3}
    \end{split}
\end{align}

\noindent $\sigma$ is the logistic sigmoid function; $h_{c_i}$, $h_{c_j} \in \mathbb{R}^d$ are the R-GCN encoded feature vectors for concepts $c_i$, $c_j$ from \cref{eq-ref1}. Each relation $r \in \mathcal{R}$ is also associated with a diagonal matrix $R_r \in \mathbb{R}^{d \times d}$.

\paragraph{Training.}
We train our graph autoencoder model using negative sampling following \citet{yang2014embedding,DBLP:conf/esws/SchlichtkrullKB18}. For triplets in $\hat{\mathcal{E'}}$ (positive samples), we create an equal number of negative samples by randomly corrupting the positive triplets. The corruption is performed by randomly modifying either one of the constituting relation or the concept.

The target label is set to $y = 1/0$ for the positive/negative triplets. In this setup, the model is then trained with standard cross-entropy loss:
\begin{equation}
\small
\begin{aligned}
  %  \begin{split}
        \mathcal{L}_{\mathcal{G'}} =-\frac{1}{2|\hat{\mathcal{E'}}|} \sum_{(c_i, r, c_j, y)\in \mathcal{T}}
        ( y\log s(c_i, r, c_j) + \\
        (1 - y)\log(1 - s(c_i, r, c_j)))
 %   \end{split}
\end{aligned}
\end{equation}
Once we train the graph autoencoder model, it will be ensured that knowledge about unseen classes (crucial during evaluation) can possibly be explained via the knowledge of seen classes and further via inter-class associations. In other words, the encoded node representations $h_{i}$ will capture commonsense information in the form of class associations and thus will be effective for the downstream ZSL task when there is the possibility of encountering unseen classes during evaluation.
\subsection{Phase 2: Commonsense Embedding Extraction \& ZSL}
\label{sec:kb_feature_extract}

\begin{figure}[t]
    \centering
    \includegraphics[width=0.8\linewidth]{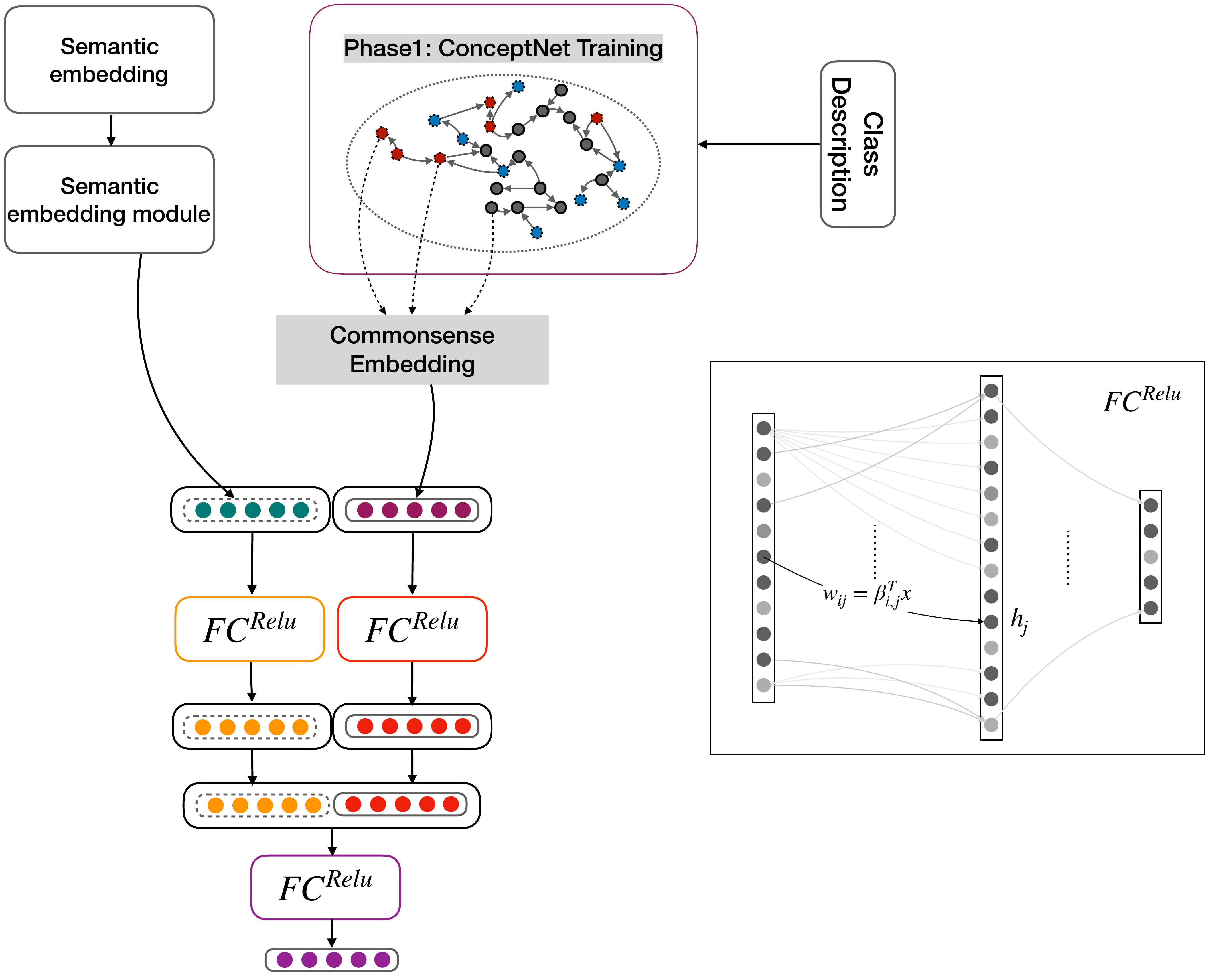}
    \caption{Fusion of CSE with semantic embedding.}
    \vspace{-0.2cm}
    \label{fig:fusion_networks}
    
\end{figure}
% \begin{figure}[h]
%     \centering
%     \includegraphics[width=\linewidth]{acl2020-templates/image_semantic_alg.pdf}
%     \caption{Image-semantics alignment networks.}
%     \label{fig:alignment_networks}
% \end{figure}
The trained graph autoencoder model from \cref{sec:kb_training}, can now be used for the commonsense embedding extraction. For a particular class name $C$, it is performed as follows:
\begin{enumerate}[leftmargin=*,noitemsep]
    \item First, we extract a subgraph from $\mathcal{G'}$, where we take all triplets which are within the vicinity of radius 2 of $C$. We call this graph $\mathcal{G'_C}$.
    \item We then make a forward pass of $\mathcal{G'_C}$ through the encoder of the pretrained graph autoencoder model. This results in feature vectors $h_j$ for all unique nodes $j$ in $\mathcal{G'_C}$. 
    \item Finally we take the feature vector $h_C$ for node $C$, to obtain the commonsense embedding $CSE_C$ for class $C$.
\end{enumerate}
We hypothesize that since unseen classes will be connected to the seen classes through the inter-class associations in $\mathcal{G'}$, $CSE_C$ will inherently capture relevant information likely to be helpful in ZSL.
\subsection{Fusion: Integrating Commonsense Embedding into the ZSL framework}
Once the commonsense embeddings for an image class-label are extracted, we need a mechanism to fuse these embeddings with the semantic embeddings that are used in the baselines. We carry out this using the following operations:
 \begin{equation}
 \small
 \begin{aligned}
        \widetilde{s_y} &= FC^{Relu}_{e}(s_y)\\
        \widetilde{c_y} &= FC^{Relu}_{f}(CSE_y)\\
        s_y' &= \widetilde{s_y} \oplus \widetilde{c_y}\\
         \end{aligned}
    \end{equation}
where $s_y$ and $CSE_y$ are the semantic and commonsense embeddings respectively. The output dimension of $FC^{Relu}_{f}$ is set equal to the size of the visual feature embedding size in the case of DeZSL.  Finally, $s_y'$ is sent to the $\Theta$ in both the baselines models-- DeZSL and RN. This can be expressed using the equations:
 \begin{equation}
 \small
 \begin{aligned}
    \Theta_{DeZSL}(s_y') &= FC^{Relu}_{b}(FC^{Relu}_{a}(s_y'))\\
    \Theta_{RN}(s_y') &= FC^{Relu}_{c}(s_y')
 \end{aligned}
 \end{equation}
 The outputs from $\Theta_{DeZSL}(s_y')$ and $\Theta_{RN}(s_y')$ are fed to respective relation module of both the baseline models (\cref{eqn:baseline1} and \cref{eqn:baseline2gamma}) -- DeZSL and RN for training.
 
 Although the adopted fusion scheme is relatively simpler than the ones proposed in recent literature, we find it very effective.  In this paper we focus only on efficacy of commonsense knowledge in ZSL rather than effectiveness of fusion methods.
\begin{table}[!t]
% \begin{center}
\small
 %\resizebox{\linewidth}{!}{
\begin{tabular}{ccc}
\hline
\textbf{Dataset} & \textbf{\#instances} & \textbf{$\mathbf{\mathcal{C}_1}$/$\mathbf{\mathcal{C}_2}$} \\
\hline\hline
\textbf{AwA1} \cite{AwA1} & 30475 & 40/10 \\
\textbf{AwA2} \cite{xian2017zero} & 37322 & 40/10 \\
\textbf{aP\&Y} \cite{farhadi2009describing} & 15339 & 20/12\\
\hline
\end{tabular}
%}
% \end{center}
\caption{\footnotesize Description of the ZSL benchmark datasets used. Note: $\mathcal{C}_1$/$\mathcal{C}_2$ - No. of \textit{seen}/\textit{unseen} classes}
\vspace{-.2cm}
\label{tab:desc}
\end{table}

\begin{table*}[ht]
\small
 \resizebox{\linewidth}{!}{
\begin{tabular}{ccc}
\hline
\textbf{\begin{tabular}[c]{@{}c@{}}Class pairs\\ \textless{}Seen, Unseen\textgreater{}\end{tabular}}                            & \textbf{\begin{tabular}[c]{@{}c@{}}Cosine\\ similarity\end{tabular}} & \textbf{\begin{tabular}[c]{@{}c@{}}Commonsense Graph\\ Associations\end{tabular}}                                                                          \\ \hline
\hline
\textless{}Aeroplane, Motorbike\textgreater{}   & 0.03                                                                 & aeroplane $\xrightarrow{\color{red}{Synonym}}$ airplane$\xrightarrow{\color{red}{RelatedTo}}$aerodynamics$\xrightarrow{\color{red}{HasContext}}$motorracing$\xrightarrow{\color{red}{RelatedTo}}$motorbike                                               \\
\textless{}Building, Statue\textgreater{}       & 0.32                                                                & building$\xrightarrow{\color{red}{RelatedTo}}$build$\xrightarrow{\color{red}{RelatedTo}}$statue                                                                                                          \\
%\textless{}Tvmonior, Diningtable \textgreater{} & 0.31                                                                 & tv\_monitor$\xrightarrow{\color{red}{Synonym}}$television monitor$\xrightarrow{\color{red}{IsA}}$monitor$\xrightarrow{\color{red}{AtLocation}}$desk$\xrightarrow{\color{red}{EtymologicallyRelatedTo}}$dish$\xrightarrow{\color{red}{Synonym}}$plate$\xrightarrow{\color{red}{RelatedTo}}$dining\_table \\
\textless{}Tvmonitor, Chair\textgreater{}       & 0.28                                                                 & tv\_monitor$\xrightarrow{\color{red}{Synonym}}$television monitor $\xrightarrow{\color{red}{IsA}}$monitor$\xrightarrow{\color{red}{AtLocation}}$desk $\xrightarrow{\color{red}{SimilarTo}}$chair                                                 \\
\textless{}Jetski, Motorbike\textgreater{}      & 0.45                                                                 & jetski$\xrightarrow{\color{red}{RelatedTo}}$motorbike                                                                                                                             \\ \hline
\end{tabular}
}
\caption{\footnotesize Examples of path between seen and unseen classes in ConceptNet.}
\vspace{-0.2cm}
\label{tab:analysis}
\end{table*}

\begin{table}[h!]
%   \centering
\small
  \resizebox{\linewidth}{!}{
   \begin{tabular}{c||c@{~~}c|c@{~~}c|c@{~~}c}
    \hline
    \multirow{2}{*}{\textbf{\begin{tabular}[c]{@{}c@{}}Model\\ Variants\end{tabular}}} & \multicolumn{2}{c|}{\bf AwA1} & \multicolumn{2}{c|}{\bf AwA2} & \multicolumn{2}{c}{\bf aP\&Y} \\
    \cline{2-7} & \bf DeZSL & \bf RN & \bf DeZSL & \bf RN & \bf DeZSL & \bf RN \\
    \hline
    \hline
  \bf HA & 68.25 & 56.47 & 65.34  & 44.12 & 25.53 & 28.90 \\
  \bf HA+CSE & \textbf{68.41} & \textbf{60.54} & \textbf{68.71} & \textbf{56.55} & \textbf{34.47} & \textbf{29.19} \\
%   \hline
  \bf HA+WNet & 67.10 & 47.14 & 60.83 & 41.3 & 20.2 & 24.0\\
  \hline
  \hline
  \bf DWE & 45.16 & 38.03 & 43.29 & 25.49 & 27.42 & 38.23 \\
  \bf DWE+CSE & \textbf{46.58} & \textbf{40.36} & \textbf{44.53} & \textbf{32.15} & \textbf{28.65} & \textbf{39.83} \\
%   \hline
    \bf DWE+WNet & 42.12 & 37.27 & 32.7 & 35.26 & 14.08 & 34.55 \\
    \hline
    \hline
  \bf HA+DWE & 66.75 & 43.38 & 66.01 & 40.88 & 20.91 & \textbf{38.45}\\
  \bf HA+DWE+CSE & \textbf{67.01} & \textbf{47.6} &  \textbf{68.52} & \textbf{46.49} & \textbf{22.35} & 37.65\\
    % \hline
  \bf HA+DWE+WNet & 65.15 & 47.38 & 63.3 & 46.27 & 17.22 & 27.53\\
  \hline
    \hline
   \end{tabular}
  }
  \caption{\footnotesize Results of model variants \& WordNet(WNet). Bold values denote best performance in a block.}
  \vspace{-0.2cm}
  \label{tab:results_main}
\end{table}

\section{Experiments}
\label{sec:experiments}
%In this section, we evaluate the effectiveness of commonsense embedding in improving the strong ZSL baselines. %More precisely, we report an ablation study to determine how addition of commonsense knowledge to semantic embedding affects ZSL performance of two baselines explained in \S \ref{sec:baseline1} and \S \ref{sec:baseline2}. 
%We compare our results with state-of-the-art methods on three popular ZSL benchmark datasets.   
% \begin{table}[h!]
% \resizebox{\linewidth}{!}{
% \begin{tabular}{c||c|cccc}
% \hline
% \multirow{2}{*}{\textbf{Datasets}} & \multirow{2}{*}{\textbf{Wang et al.}} & \multicolumn{2}{c}{\textbf{HA+CSE}} & \multicolumn{2}{c}{\textbf{DWE+CSE}} \\ \cline{3-6} 
%  &  & \textbf{DeZSL} & \textbf{RN} & \textbf{DeZSL} & \textbf{RN} \\ \hline
% \textbf{AwA1} & 34.55  &\textbf{68.41} & 65.54 & 47.58 & 41.36 \\ %15.50
% \textbf{AwA2} & 34.59 & \textbf{70.31} & 56.55 & 44.53 & 42.15\\
% \textbf{aP\&Y} & 14.76 & \textbf{44.47} & 29.19 & 28.65 & 40.23\\ \hline
% \end{tabular}
% }
% \caption{\footnotesize Comparison with \citet{wang2018zero}}
% \vspace{-0.2cm}
% \label{tab:results_sota}
% \end{table}
\begin{table}[h!]
\resizebox{\linewidth}{!}{
\begin{tabular}{c||ccc|cccc}
\hline
\multirow{2}{*}{\textbf{Datasets}} & \multirow{2}{*}{\textbf{W\tiny{ang et. al}}} &\multirow{2}{*}{\textbf{W\tiny{ang et. al} \normalsize{ + HA}}} & 
\multirow{2}{*}{\textbf{W\tiny{ang et. al}\normalsize{ + DWE}}} &
\multicolumn{2}{c}{\textbf{HA+CSE}} & \multicolumn{2}{c}{\textbf{DWE+CSE}} \\ \cline{5-6} \cline{7-8}
 &  & & & \textbf{DeZSL} & \textbf{RN} & \textbf{DeZSL} & \textbf{RN} \\ \hline
\textbf{AwA1} & 34.55 & 46.96 & 41.40  &\textbf{68.41} & 65.54 & 47.58 & 41.36 \\ %15.50
\textbf{AwA2} & 34.59 & 47.25 & 43.34 & \textbf{68.71} & 56.55 & 44.53 & 42.15\\
\textbf{aP\&Y} & 14.76 & 34.50 & 33.19 & \textbf{34.47} & 29.19 & 28.65 & 40.23\\ \hline
\end{tabular}
}
\caption{\footnotesize Comparison with \citet{wang2018zero}}
\vspace{-0.2cm}
\label{tab:results_sota}
\end{table}
% \vspace{-1em}
\subsection{Experimental Setup}
We validate our commonsense embedding by performing experiments on three benchmark datasets on the task of ZSL. We choose two mid-sized datasets namely Animals with Attributes1 (\textbf{AwA1}) \cite{AwA1} and Animals with Attributes2 (\textbf{AwA2}) \cite{xian2017zero}. They share the same animal classes (such as horse, zebra, sheep, blue-whale etc.) but different images. aPascal \& aYahoo (\textbf{aP\&Y}) \cite{farhadi2009describing} is small dataset containing image-classes form varied domains (such as dogs, aeroplane, jet-ski, potted-plant etc.). Details regarding these datasets are provided in \cref{tab:desc}. We adopt GBU settings \cite{xian2017zero} commonly used in ZSL literature. We use ConceptNet 5.5 \cite{speer2017conceptnet} in all our experiments.
\subsection{Features Used}
\label{sec:features_used}
We use the following features in our ZSL baselines.

\noindent \textbf{Visual Embedding}: As used in \cite{xian2017zero,zhang2017learning}, our visual embeddings are 2048-dim top-layer pooling units of the
101-layered ResNet. It is pre-trained on ImageNet 1K \cite{deng2009imagenet} and not fine-tuned.

\noindent \textbf{Semantic Embeddings}:
As discussed in the earlier sections, we use two different kinds of semantic embeddings:
\begin{itemize}[leftmargin =*,noitemsep]
    \item \textbf{Distributed Word Embeddings (DWE):} We use word2vec vectors pre-trained on part of Google News dataset \cite{mikolov2013efficient} to cast each class name into a 300-dimensional vector representation.
    \item \textbf{Human Defined Attributes}: Human defined attributes (such as \emph{furry,stripes}) are 85 dimensional attribute vectors for Awa1 and AwA2. For aP\&Y, we have 64 dimensional attribute vectors.
\end{itemize}
\subsection{Implementation Details}
%All models and related codes in this work are written using python using pytorch deep learning framework \cite{paszke2017automatic}. 
In both the baselines, number of nodes in FC layers are chosen from --- 512, 1024, 2048 and 4096 and are empirically determined using 10\% data from the training set validaiton set. Both baselines are trained using ADAM optimizer \cite{kingma2014adam} with a learning rate of $10^{-5}$.

\subsection{Model Variants and Baselines}
\label{sec:modelvar}
In order to study the effect of commonsense knowledge, we look into the following variants:
\begin{itemize}[leftmargin =*,noitemsep]
    \item \textbf{HA}: Only using human defined attributes.
    \item \textbf{DWE}: Only constituting of word2vec DWEs.
    \item \textbf{HA+DWE}: Fusion of HA + DWE (as in \cref{fig:fusion_networks}). 
    \item  \textbf{HA+CSE}: We fuse human defined attributes with commonsense embedding.
    \item \textbf{DWE+CSE}: Word2vec is fused with commonsense embedding in order to compare against performance of DWE. 
    \item \textbf{HA+DWE+CSE}: We fuse CSE with the fusion of HA and DWE for comparison with HA+DWE.
    \item \textbf{\citet{wang2018zero}}: We compare our results with \cite{wang2018zero} as their approach is closest to ours. The purpose of this experiment is to check whether the existing knowledge graph-based approaches can produce results similar to ours. \citet{wang2018zero} use a knowledge graph, NELL, and find the shortest path between each class pair in it. This process generates a subgraph. They apply a GCN on this subgraph to compute the node embeddings. As some of these nodes actually represent the classes in the dataset, 
% The output from the multilayer GCN is compared against the image features of the corresponding class labels using MSE loss. Once the framework is trained, the class embeddings can be obtained from the network. Similar to how we use commonsense features in our two ZSL baselines, we feed these embeddings into the ZSL baselines.
they then feed these class embeddings into ZSL. We re-implemented their method by replacing NELL with ConceptNet. Note that the only similarity between \citet{wang2018zero} and our work is that we both use graph convolution, but the methodologies and motivation are completely different. \citet{wang2018zero} use GCN on a knowledge graph to transform initial pre-trained word embeddings and find an association among the transformed embeddings to visual embedding. However, we extract sub-graphs for all seen and unseen classes from the knowledge graph and pass them through a graph convolutional autoencoder model to extract a commonsense embedding. As such, our results are superior to \citet{wang2018zero}'s as indicated in \cref{tab:results_sota}. 
\end{itemize}

\subsection{Evaluation Scheme}
We follow conventional ZSL setting -- training and testing classes are \textbf{completely} disjoint. This setting is standard for majority of ZSL works \cite{wang2019survey} including the original papers we formulate our baselines upon \cite{zhan-etal-2017-network,sung2018learning}. All experiments report \textbf{top-1} accuracy.
\subsection{Results and Analysis}
We feed commonsense embedding (CSE) into two strong baseline models as discussed in \S \ref{sec:prelem}. To do so, we fuse commonsense CSE with two different semantic embeddings 1) human defined attributes (HA); 2) word2vec (DWE). Experimental results are shown in \cref{tab:results_main}. \par
We observe a common trend of increase in ZSL recognition accuracies when CSE is fused with HA, DWE as well as HA+DWE. This demonstrates the role that commonsense knowledge can play in discovering meaningful associations between seen and unseen classes, which in turn facilitates knowledge transfer. We notice that the overall performance of the models on aP\&Y is lesser compared to the other two datasets. We believe this happens because the seen and unseen class labels in the $aP\&Y$ dataset are from varied domains e.g., \emph{bird, statue, train}. Wherein, in the other two datasets the class labels are from only \emph{animal} domain. 

\paragraph{Comparison with WordNet:} Can we replicate our performance with a knowledge graph that does not contain commonsense information? To answer this, we replaced ConceptNet with WordNet(WNet) and applied our method. Wordnet~\cite{fellbaum2010wordnet} is a lexical knowledge graph that contains semantic connections such as \emph{synonym, hypernym, hyponym, antonym} among the words. WordNet does not contain commonsense related information that we see in ConceptNet. Performances consistently drop by a large margin when WordNet is replaced with ConceptNet (\cref{tab:results_main}). % We conjecture that the performance drops because WordNet lacks the richer commonsense knowledge that is available in ConceptNet.

\paragraph{Effect of Commonsense Knowledge:} We take a qualitative look at the effectiveness of commonsense knowledge in \cref{tab:analysis}. Here, Column 1 indicates a pair of  \textless{}\textit{seen}, \textit{unseen}\textgreater{} classes. Column 2 shows cosine similarity of their word2vec class embeddings. Column 3 reports association of these classes in commonsense graph. Although the pairs are not contextually similar as reflected in cosine similarity between their word2vec embeddings, they still have a path of associations between them. Even in cases of visually similar classes such as \textless{} Jetski, Motorbike \textgreater{} we see low cosine similarity of 0.45. %indicting word2vec does not provide much help in knowledge transfer from 'Jetski' to 'Motorbike'.
Whereas in our commonsense graph, they have a direct relation ('relatedTo') -- jetski$\xrightarrow{\color{red}{RelatedTo}}$motorbike. %This means due to message passing during training of GCN, Common Sense Embedding (CSE) of 'Jetski' will carry a great deal of information about the class 'Motorbike'. This reflects in the ZSL performances and justifies our claim as well.  
This ensures that commonsense embedding of 'Motorbike' can carry a great deal of information about the embedding of 'Jetski' because neighborhood message passing is the underlying formulation of the trained GCN model. Our corresponding ZSL performance gain also corroborates with this. 
% \vspace{-0.3cm}
\paragraph{Comparison with \citet{wang2018zero}:} 

% We have applied this architecture on ConceptNet and the results are shown in \cref{tab:e2ekg}. 
The results of this experiment are shown in \cref{tab:results_sota}. It can be seen that our proposed approach outperforms \citet{wang2018zero} by a significant margin. It shows that our knowledge graph encoding method and its usage in the ZSL baselines are superior.
\section{Conclusion}
\label{sec:conclusion}
In this work, we built upon ConceptNet to extract commonsense embeddings 
% which we used 
to address the task of Zero Shot Learning (ZSL). To generate commonsense embeddings, we applied Graph Convolutional Network on the ConceptNet commonsense graph. Experimental results showed that our method achieved significant performance improvement over two strong baselines on three well known datasets. %In future, we plan to extend our framework to generalized ZSL setting and explore to what extent the commonsense embedding improves if we merge multiple knowledge graphs in a unified knowledge representation.

% We will make our code publicly available.

%extend this work by improving the commonsense embedding generation method. It would be also interesting to explore to what extent the commonsense embedding improves if we merge multiple knowledge graphs in a unified knowledge representation.
\bibliography{anthology,emnlp2020}
\bibliographystyle{acl_natbib}
\end{document}